\def\year{2017}\relax
\documentclass[letterpaper]{article} 
\usepackage{aaai}  
\usepackage{times}  
\usepackage{helvet}  
\usepackage{courier}  
\usepackage{url}  
\usepackage{graphicx}  
\usepackage{xcolor}
\usepackage{subfig}

\usepackage{titlesec}
\titlespacing*{\paragraph} {0pt}{0.8ex plus 1ex minus .2ex}{1em}

\newcommand{\nlstring}[1]{{\em #1}}

\begin{document}

%
\title{Visual Reasoning with Natural Language}

\author{Stephanie Zhou\thanks{Contributed equally.}, Alane Suhr\footnotemark[1], and Yoav Artzi \\
Dept. of Computer Science and Cornell Tech\\
Cornell University\\
New York, NY 10044\\
{\tt sz244@cornell.edu suhr@cs.cornell.edu yoav@cs.cornell.edu}}

\maketitle

\section{Introduction}

\begin{figure}[!tbp]
\centering
\frame{\includegraphics[width=0.5\linewidth]{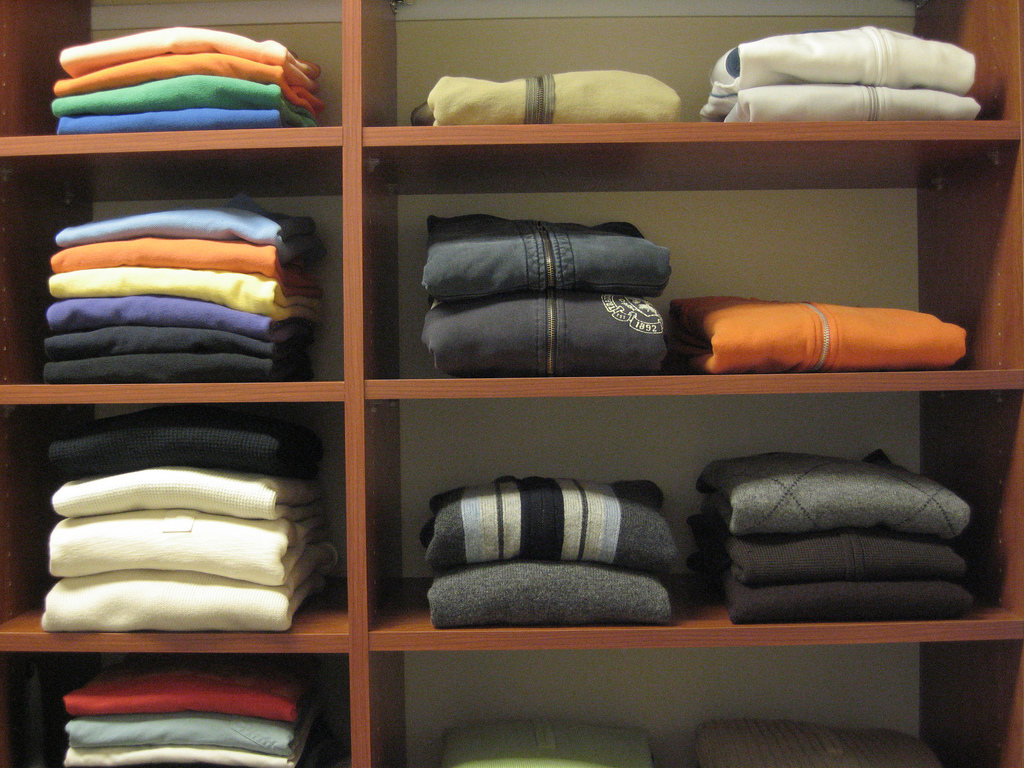}} \\[5pt]
\fbox{\begin{minipage}{0.95\linewidth}
\nlstring{Fold and place the sweatshirt on the top shelf, and make sure the stacks are the same color and evenly distributed.}
\end{minipage}}
		\vspace{-8pt}
\caption{An example instruction that may be given to a household robot.}
		\vspace{-10pt}
\label{fig:realexample}
\end{figure}

Natural language provides a widely  accessible and expressive interface for robotic agents. 
To understand language in complex environments, agents must reason about the full range of language inputs and their correspondence to the world. 
For example, consider the scenario and instruction in Figure~\ref{fig:realexample}. 
To execute the instruction, the robot must  identify \nlstring{the top shelf}, recognize the two \nlstring{stacks} as sets of items, compare items, and reason about the content and size of the sets. 
Such reasoning over language and vision is an open problem that is receiving increasing attention~\cite{Antol:15vqa,Chen:15coco,Johnson:16clevr}. 
While existing data sets focus on visual diversity, they do not display the full range of natural language expressions, such as counting, set reasoning, and comparisons. 

We propose a simple natural language visual reasoning task, where the goal is to predict if a descriptive statement paired with an image is true for the image. 
This abstract describes our existing synthetic images corpus~\cite{Suhr:17visual-reason} and  current work on collecting real vision data.

\section{Related Work}

Several tasks focus on language understanding in visual contexts, including caption generation~\cite{Chen:15coco,Young:14flickr30k,Plummer:15flickr30k}, visual question answering~\cite{Antol:15vqa}, referring expression resolution~\cite{Matuszek:12,Krishnamurthy:13} and generation~\cite{Mitchell:10,FitzGerald:13}, and mapping of instructions to actions~\cite{MacMahon:06,Chen:11,Artzi:13,Bisk:16nl-robots,Misra:17instructions}. 
We focus on visual reasoning with emphasis on linguistic diversity. 
The most related resource to ours is CLEVR~\cite{Johnson:16clevr}, where questions are paired with synthetic images. 
However, in contrast to our work, both language and images are synthetic.

\begin{figure}[t]
	\centering
	\fbox{
		\begin{minipage}{0.95\linewidth}
			\centering
			\begin{footnotesize}
			\frame{\includegraphics[width=0.5\linewidth]{example1-shapes}}  \\
			\textit{There is a box with 2 triangles of same color nearly touching each other.}
			\end{footnotesize}
		\end{minipage}
	}\\[1ex]
	\fbox{
		\begin{minipage}{0.95\linewidth}
			\centering
			\frame{\includegraphics[width=0.5\linewidth]{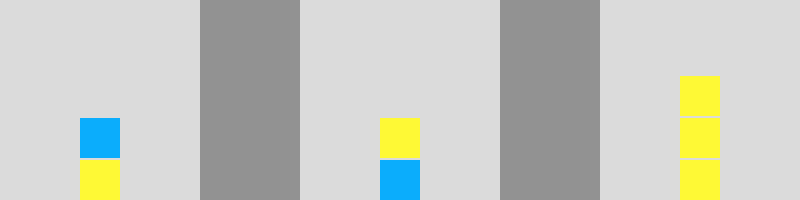}} \\
			\begin{footnotesize}
			\textit{There are two towers with the same height  but their base is not the same in color.}
			\end{footnotesize}
		\end{minipage}
	}
		\vspace{-10pt}
		\caption{Example for natural language visual reasoning. The top sentence is false, while the bottom is true.}
		\vspace{-15pt}
	\label{fig:examples}
\end{figure}

\section{Task}
Given an image and a natural language statement, the task  is to predict whether the statement is true in regard to the image. 
Figure~\ref{fig:examples} shows two examples with generated images. The statement in the top example is true in regard to the given image, while the lower example is false. 
We evaluate system performance using accuracy. 
This provides a straightforward evaluation metric, in contrast to other related tasks, which use partial credit metrics, such as \textsc{Bleu}.

\section{Synthetic Image Data}

\begin{figure}[t]
\centering
\fbox{\begin{minipage}{0.97\linewidth}
	\centering
		\begin{minipage}{0.95\linewidth}
			\begin{center}
			\renewcommand{\arraystretch}{2.1}
			\begin{tabular}{cc}
				(A) & \raisebox{-.5\height}{\frame{\includegraphics[width=0.4\linewidth]{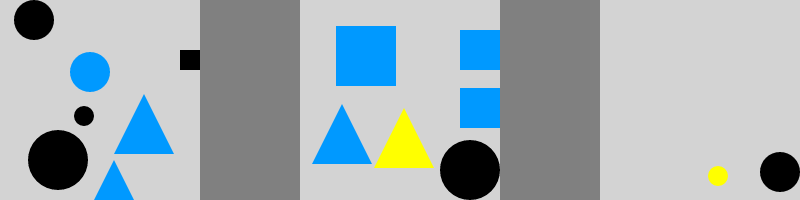}}} \\[1ex]
				(B) & \raisebox{-.5\height}{\frame{\includegraphics[width=0.4\linewidth]{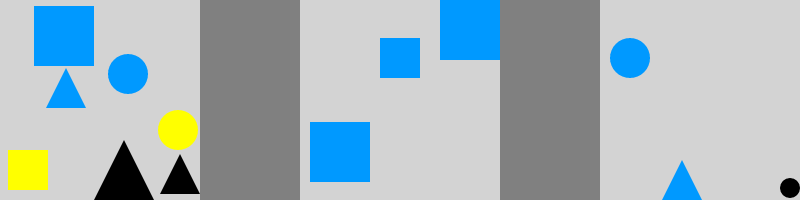}}} \\[1ex]
				(C) & \raisebox{-.5\height}{\frame{\includegraphics[width=0.4\linewidth]{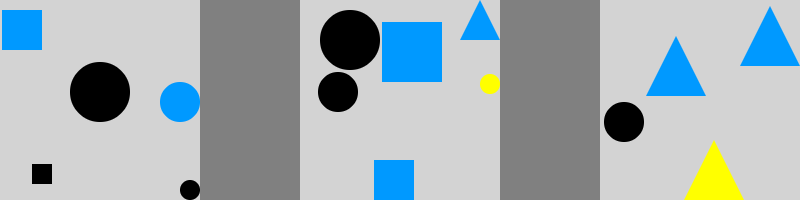}}} \\[1ex]
				(D) & \raisebox{-.5\height}{\frame{\includegraphics[width=0.4\linewidth]{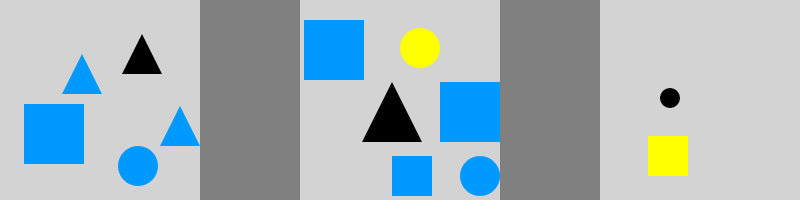}}} \\
			\end{tabular}
			\end{center}
			
			\begin{footnotesize}
			Write one sentence. This sentence must meet all of the following requirements: 
			\begin{itemize}
			\item It describes A.
			\item It describes B.
			\item It does \emph{not} describe C.
			\item It does \emph{not} describe D.
			\item It does \emph{not} mention the images explicitly (e.g. ``In image A, ...'').
			\item It does \emph{not} mention the order of the light grey squares (e.g. ``In the rightmost square...'')
			\end{itemize}
			There is no one correct sentence for this image. There may be multiple sentences which satisfy the above requirements. If you can think of more than one sentence, submit only one. 
	
			\end{footnotesize}
		\end{minipage}
		\end{minipage}}
				\vspace{-10pt}
	\caption{Sentence writing prompt. The top sentence in Figure~\ref{fig:examples} was generated from this prompt.}
			\vspace{-15pt}
	\label{fig:stage_1}
\end{figure}

In \citeauthor{Suhr:17visual-reason}~\shortcite{Suhr:17visual-reason}, we present the Cornell Natural Language Visual Reasoning (NLVR) corpus. 
The corpus includes statements paired with synthetic images.
Using synthetic images enables control of the visual content and reasoning required to distinguish between images. 
We briefly review the data, collection process, and considerations, and refer to the original publication for the details.

\paragraph{Data Collection}
We define a two-stage process: sentence writing and validation. 
Figure~\ref{fig:stage_1} illustrates the sentence writing stage. 
This task requires workers to identify similarities and differences between images, and requires careful reasoning, which  is reflected in the collected language. 
We show workers four generated images, each made of three boxes containing shapes. 
The first two images are generated independently. 
The third and fourth are generated from the first and second by shuffling objects. 
This discourages trivial sentences, such as \nlstring{there is a blue triangle}. 
We ask for a sentence that is true for the first two images, and false for the others. 
We instruct workers that sentences may not refer to the order of boxes. 
This enables permuting the boxes while retaining the statement truth value. 
We pair each image with the written sentence to create four pairs. 
In the validation stage, we ask for a label for each pair. 
While the truth-value can be inferred from the sentence-writing stage,  validation increases data quality. 
Finally, we generate six image-sentence pairs by permuting the three boxes in each image.

\paragraph{Data Statistics}

We collect 3,962 unique sentences for a total of 92,244 sentence-image pairs. 
We create four sets: 80.7\% for training, 6.4\% for development, and the rest for two test sets. 
One test set is public, and the second is unreleased and used for the task leaderboard. 
For testing and development sets we collect five validation judgements for each pair, and observe high inter-annotation agreement (Krippendorff's $\alpha = 0.31$ and Fleiss' $\kappa = 0.808$).

\paragraph{Analysis}

We analyze our corpus and existing corpora for linguistic complexity. 
We classify for a broad set of linguistic phenomena, including quantification, cardinal reasoning, syntactic ambiguity, and semantic and pragmatic features (e.g., coreference and spatial relations). 
The details of the analysis are in \citeauthor{Suhr:17visual-reason}~\shortcite{Suhr:17visual-reason}. 
Our analysis shows our data is significantly more linguistically diverse than VQA.
For example, 66\% of our sentences refer to exact counts, whereas this occurs in only 12\% of sentences in VQA. 

\paragraph{Baselines}

We evaluate the difficulty of the task with multiple baselines. 
We construct two models that use only one of the input modalities to measure biases. 
Both perform similarly to the majority-class baselines. 
The best-performing model is neural module networks~\cite{Andreas:16nmn}, which achieves  62\% on the unreleased test set. 
The original publication includes a break down  of sampled errors per analyzed linguistic phenomena.

\section{Real Vision Data}

\begin{figure}
	\centering
	\fbox{
		\begin{minipage}{0.95\linewidth}
			\centering
			\frame{\colorbox{gray!20}{\includegraphics[height=12ex]{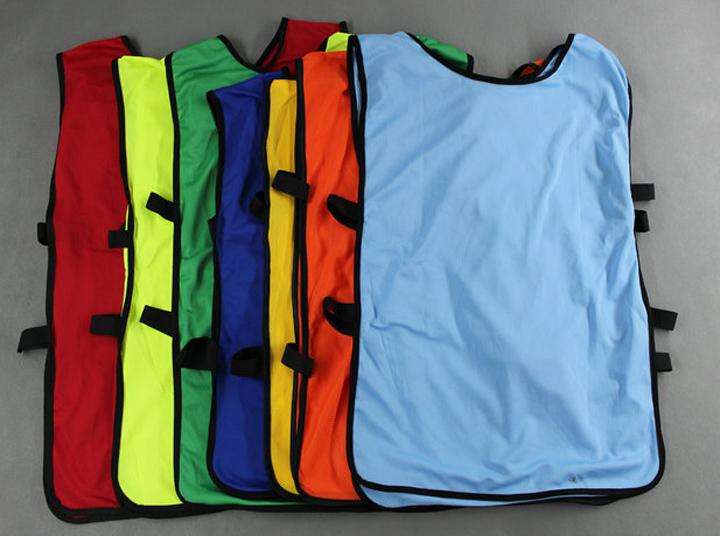}~{\includegraphics[height=12ex]{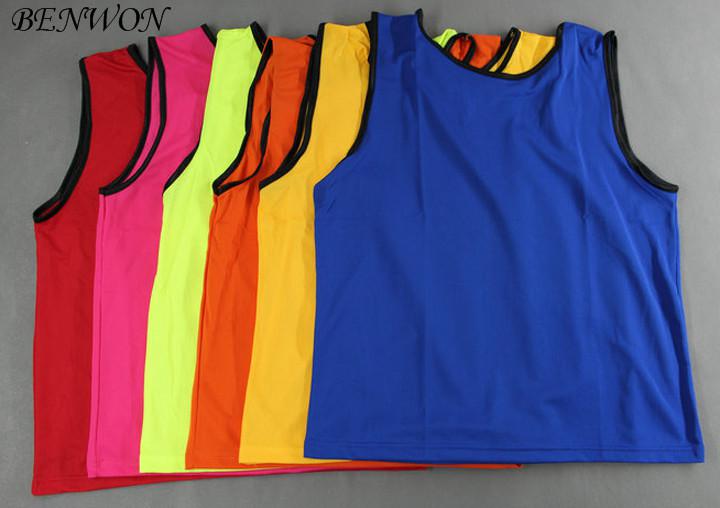}}}} \\
			\begin{footnotesize}
			\textit{A red vest is furthest to the left in at least one paired image.}
			\end{footnotesize}
		\end{minipage}
	}\\
	\fbox{
		\begin{minipage}{0.95\linewidth}
			\centering
			\begin{footnotesize}
			\frame{\colorbox{gray!20}{\includegraphics[height=12ex]{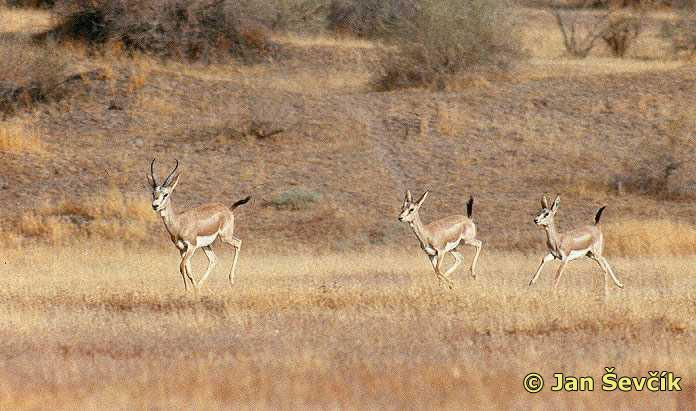}~{\includegraphics[height=12ex]{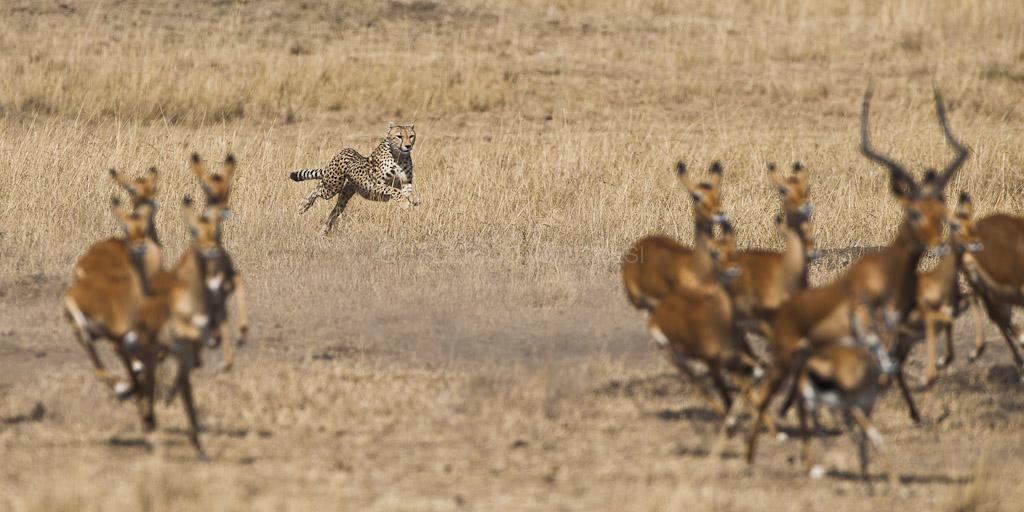}}}}  \\
			\textit{The gazelle in both pictures are running the same direction}
			\end{footnotesize}
		\end{minipage}
	}
			\vspace{-10pt}
		\caption{Crowdsourced sentences and images from our ongoing work. The truth value of the top sentence is true, while the bottom is false.}
		\vspace{-15pt}
	\label{fig:real}
\end{figure}

We are currently collecting an NLVR real vision data set. 
Our goal is to collect statements displaying a variety of linguistic phenomena, such as counting, spatial relations, and comparisons. 
In contrast to our use of synthetic images, we aim for realistic visual input, including a broad set of object types and scenes. 
Figure~\ref{fig:real} shows initial examples. 
To correctly reason about the top statement, the system must maximize a spatial property and identify the number of images in which it holds. 
To understand the second statement, the agent has to consider several unique objects and compare a certain property they all demonstrate--the direction they face.

\section{Conclusion}

We describe a task for language and vision reasoning, and a newly released vision and language data set, the Cornell Natural Language Visual Reasoning (NLVR) corpus.  
While the task is straightforward to evaluate, it requires complex reasoning. 
Performance of existing methods demonstrates the challenge the data presents. 
We also discuss ongoing work on collecting similar data that includes both linguistically diverse text and real vision challenges.

\bibliography{main}

\end{document}